\documentclass[conference]{IEEEtran}
\IEEEoverridecommandlockouts
\usepackage[numbers]{natbib}
\usepackage{cite}
\usepackage{amsmath,amssymb,amsfonts}
\usepackage{algorithmic}
\usepackage{graphicx}
\usepackage{float}
\usepackage{textcomp}
\usepackage{booktabs}
\usepackage{xcolor}
\usepackage{multirow}
 \usepackage{url}
 \usepackage{caption}
\usepackage{authblk}
\usepackage[utf8]{inputenc}
\def\BibTeX{{\rm B\kern-.05em{\sc i\kern-.025em b}\kern-.08em
    T\kern-.1667em\lower.7ex\hbox{E}\kern-.125emX}}
\begin{document}

\title{\fontsize{16}{20}\selectfont Towards Patronizing and Condescending Language in Chinese Videos: A Multimodal Dataset and Detector\\

}

\author[1]{\textit{Hongbo Wang}}
\author[1]{\textit{Junyu Lu}}
\author[2]{\textit{Yan Han}}
\author[1]{\textit{Kai Ma}}
\author[1]{\textit{Liang Yang}}
\author[1]{\textit{Hongfei Lin}}

\affil[1]{Dalian University of Technology, China} \affil[2]{University of Tsukuba, Japan}
%\author[1]{\textit{Hongbo Wang}}
%\author[1]{\textit{Junyu Lu}}
%\author[2]{\textit{Yan Han}}
%\author[1]{\textit{Liang Yang}}
%\author[1]{\textit{Hongfei Lin}}

%\affil[1]{Dalian University of Technology}
%\affil[2]{Tsukuba University}

\renewcommand\Authands{ and }
\maketitle

\begin{abstract}
Patronizing and Condescending Language (PCL) is a form of discriminatory toxic speech targeting vulnerable groups,
threatening both online and offline safety. While toxic speech research has mainly focused on overt toxicity, such as hate speech, microaggressions in the form of PCL remain underexplored. Additionally, dominant groups’ discriminatory facial expressions and attitudes toward vulnerable communities can be more impactful than verbal cues, yet these frame features are often overlooked. In this paper, we introduce the PCLMM dataset, the first Chinese multimodal dataset for PCL, consisting of 715 annotated videos from Bilibili, with high-quality PCL facial frame spans. We also propose the MultiPCL detector, featuring a facial expression detection module for PCL recognition, demonstrating the effectiveness of modality complementarity in this challenging task. Our work makes an important contribution to advancing microaggression detection within the domain of toxic speech.
\end{abstract}

\begin{IEEEkeywords}
Patronizing and Condescending Language, Toxic Speech, Multimodal, Video, Facial Expression.
\end{IEEEkeywords}

\section{Introduction}
The rapid development of social media has exceeded expectations. Since around 2010, self-media has gradually gained prominence in disseminating ideas on mainstream platforms, such as the English platforms YouTube, TikTok \citep{newman2021reuters}, and the Chinese platform Bilibili \citep{peng2023exploring}. While these video platforms have created significant economic benefits and influence, they have also accelerated the spread of harmful content. Despite the robust regulations enforced by mainstream online platforms to reduce the risk of dangerous content, these measures primarily slow the dissemination of videos with clearly aggressive content, such as hate speech \citep{mathew2021hatexplain}, but overlook microaggressions targeting vulnerable communities, known as Patronizing and Condescending Language (PCL) \citep{perez2020don}.

PCL is a form of discriminatory toxic speech targeting vulnerable groups, such as individuals with disabilities, children, and the elderly, reflecting a superior attitude towards these communities \citep{perez2020don}. Although the construction of PCL corpora has advanced \citep{perez2020don,wang2019talkdown,wang2023ccpc} and the researchers also established specialized evaluation tracks, utilizing various improved deep learning networks to advance related research \citep{perez2022semeval, deng2022beike, lu2022guts}, current PCL research remains text-based. Unlike traditional toxic speech, such as hate speech, PCL lacks explicit offensive words, making it more subtle and implicit \citep{ng2007language}. The characteristics of PCL suggest that integrating multimodal approaches, especially discriminatory facial expressions, will contribute to breakthroughs in this field. Although multimodal frameworks have begun to emerge in hate detection \citep{das2023hatemm, maity2024toxvidlm}, this direction remains unexplored for PCL. Moreover, current research is limited to English and lacks attention to vulnerable groups in other language communities. 

In this paper, we introduce a multimodal dataset and corresponding detector designed to enhance the automated detection of microaggressions on video platforms, aiming to protect vulnerable communities. We introduce PCLMM, the first multimodal dataset for detecting PCL in videos, comprising 715 annotated videos, over 21 hours of content from Bilibili, one of China's largest video platforms. PCLMM includes a wide range of vulnerable communities in China and is publicly available to support further research.\footnote{The dataset and code for this project have been open-sourced at https://github.com/dut-laowang/PCLMM.} We also propose the MultiPCL Detector, which integrates facial expression features with video, text, and audio to enhance the detection of discriminatory language. Our research focuses on the Chinese context due to the prevalence of its vulnerable groups, and our findings are also relevant to English-speaking contexts. Our contributions are summarized as follows: (1) We develop and release PCLMM, the first multimodal PCL dataset, including 715 Bilibili videos (21+ hours) annotated as patronizing (PCL) or non-patronizing (non-PCL), with annotated PCL facial frame spans. (2) We introduce the MultiPCL detector, which integrates facial expressions, video, text, and audio, demonstrating significant improvements in detection accuracy. (3) Our sentiment and toxicity analysis indicates that PCL possesses a certain level of ambiguity, and our detector can effectively identify these marginal features.

\begin{figure*}[h] % 使用 figure* 环境来跨越双栏
\centering
\includegraphics[width=\textwidth]{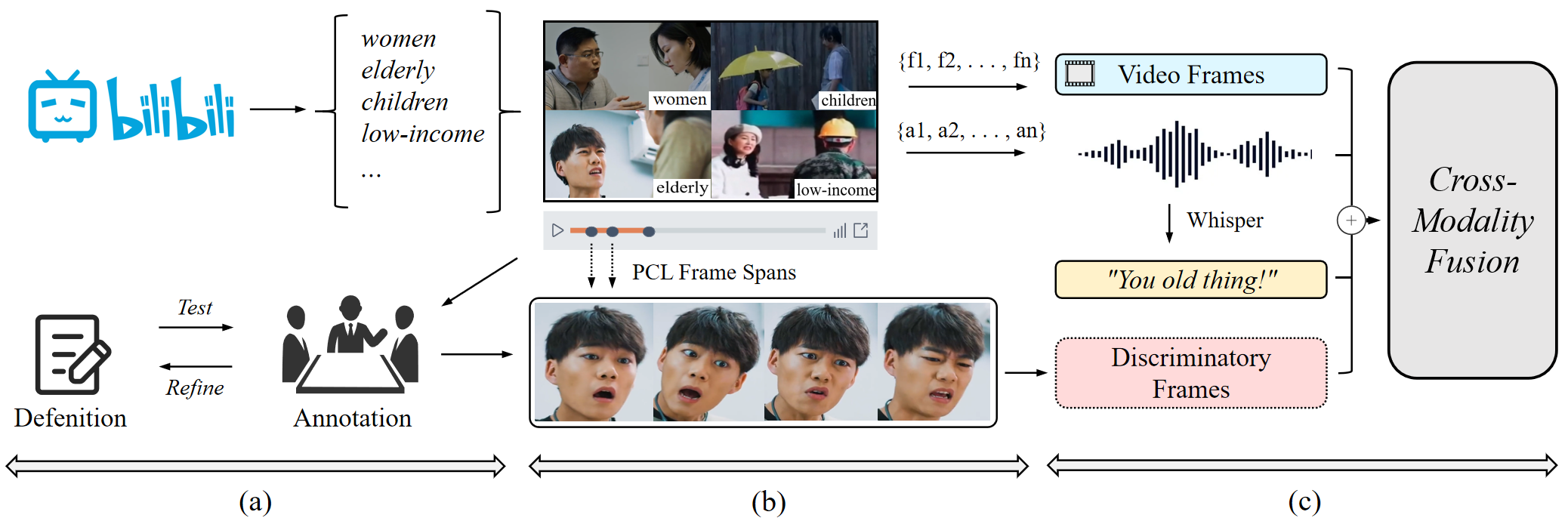} % 将图片宽度设为双栏宽度
\caption{The presentation of our multimodal PCL framework: (a) Data Collection. Refining annotation guidelines and gathering data from Bilibili. (b) PCLMM dataset. A high-quality annotated dataset with PCL frame spans. (c) MultiPCL detector. Utilizing a cross-attention mechanism to extract and integrate features from facial expressions, video, text, and audio modalities. }
\label{fig:figure1}
\end{figure*}

\section{PCLMM Dataset}
\subsection{Overview}
In this section, we outline the construction of the PCLMM dataset. We developed a comprehensive semantic definition of PCL in Chinese to create annotation guidelines. Using six key vulnerable community categories from the Chinese internet, we compiled a keyword list and collected videos via targeted searches. The dataset was manually annotated by three annotators, followed by sentiment and toxicity analysis.
\subsection{Definition Development}\label{sec:3.2}
PCL typically targets vulnerable groups, but this definition often doesn't align with the Chinese context, where the concept of vulnerable groups differs from that in English-speaking communities. For example, PCL toward immigrants is rare in China due to policy reasons. Building on \citep{perez2020don, wang2023ccpc}, we proposed a comprehensive definition of PCL tailored to the Chinese context, serving as our annotation guide.

\textit{Chinese PCL refers to discriminatory, falsely sympathetic, and hypocritical remarks directed at six vulnerable groups within the Chinese community: disabled individuals, women, the elderly, children, single-parent families, and low-income groups. A key feature of PCL is the speaker's condescending attitude, making statements that do not improve the group's situation. PCL expressions are often accompanied by contemptuous and discriminatory facial expressions.}
To minimize subjective discrepancies, we specified the following cases to be annotated as non-PCL:
\begin{itemize}
\item Vulnerable individuals describing their own experiences of unfair treatment.
\item Objective news reports on discriminatory incidents.
\item Public service announcements containing discriminatory content but lacking discriminatory intent.
\end{itemize}
\subsection{Data Collection}
Based on \ref{sec:3.2}, we identified six major vulnerable communities on the Chinese internet. We expanded each community into a list of 10 commonly used keywords and designed a lexicon of offensive and discriminatory terms as query keys. These were matched with the keyword list to generate the final search set (e.g., adding the query \textit{discrimination} to the keyword \textit{elderly care}). Our search list included 1800 keyword-value pairs, retrieving 2654 preliminary videos. We retained videos ranging from 30 seconds to 5 minutes and filtered out damaged and irrelevant videos, and we also used masking techniques to obscure all possible watermarks that might disclose user privacy. Finally, we got 715 high-quality annotable samples.
%\footnote{The complete keyword list can be found at https://tomasjwyu.github.io/AutoPrepDemo/}
\begin{table}[H]
\caption{\textnormal{Statistics of PCLMM. \textit{PCL Frame Spans} refer to the statistics of the patronizing spans within PCL videos, and $\mu$ represents the average value.}}
\begin{center}
\begin{tabular}{c|c|c|c|c}
\toprule[1.2pt]  % 加粗的上边线
\textbf{} & \textbf{\textit{Non-PCL}} & \textbf{\textit{PCL}} & \textbf{\textit{PCL Frame Spans}} & \textbf{\textit{Total}} \\
\midrule
Total num& 519 & 196 & 330 &715  \\
Total len (hrs) & 15.1 & 6.5 & 2.3 & 21.6 \\
Total frame (M) &  1.6& 0.7 & 0.2 & 2.3 \\
$\mu$ Video len (min) & 1.7 & 1.9 & 0.4 & 1.8 \\
$\mu$ Text len (char)& 455 & 536 & 158 & 477 \\
\bottomrule[1.2pt]  % 加粗的下边线
\end{tabular}
\label{tab1}
\end{center}
\vspace{-5mm}
\end{table}
 \subsection{Data Annotation}
Two trained Ph.D. students annotated the videos, with a third as the reviewer (two males, one female, aged 25-30, all in computer science and focused on toxicity detection). Annotators were compensated based on the number of annotations completed. Videos were labeled as PCL or non-PCL following \ref{sec:3.2}. An initial set of 30 videos (20 non-PCL, 10 PCL) was used to reach consensus on discrepancies. To minimize harm, annotators were limited to 20 videos per day and reported their psychological state. The CVAT tool \citep{cvat} further recorded facial expression spans for PCL videos, while non-PCL videos had no patronizing spans. Fleiss’ Kappa \citep{fleiss1971measuring} measured inter-annotator agreement (IAA = 0.72), with the third annotator resolving discrepancies. Finally, we obtained 196 PCL and 519 non-PCL videos. 
\vspace{-2mm}
\subsection{Data Statistics}
The PCLMM dataset contains 715 videos, totaling 21 hours of content, with an average video length of 1.80 mins and a frame rate of 30 FPS, comprising 2.3M frames. Approximately 27.4\% of the videos were labeled as patronizing, aligning with the distribution of PCL data on internet platforms. Detailed dataset statistics are shown in Table \ref{tab1}.
\subsection{Data Analysis}
\subsubsection{Sentiment Analysis}
We used the advanced open-source model DeepFace \citep{serengil2024lightface} to analyze facial expressions in the PCLMM dataset. We sampled 20 videos per community from both PCL and non-PCL subsets, totaling 240 samples. For PCL, 10 facial expressions were selected from annotated PCL frame spans; for non-PCL, 10 expressions were from general frames. As shown in Figure \ref{fig:figure1}, non-PCL expressions were predominantly positive or neutral, while PCL expressions conveyed more negative emotions such as anger, sadness, and disgust. Some PCL expressions were misclassified as 'happy,' despite indicating superiority and contempt, highlighting the limitations of basic positive-negative classification in detecting PCL.
\begin{figure}[h]
\centering
\includegraphics[width=0.9\columnwidth]{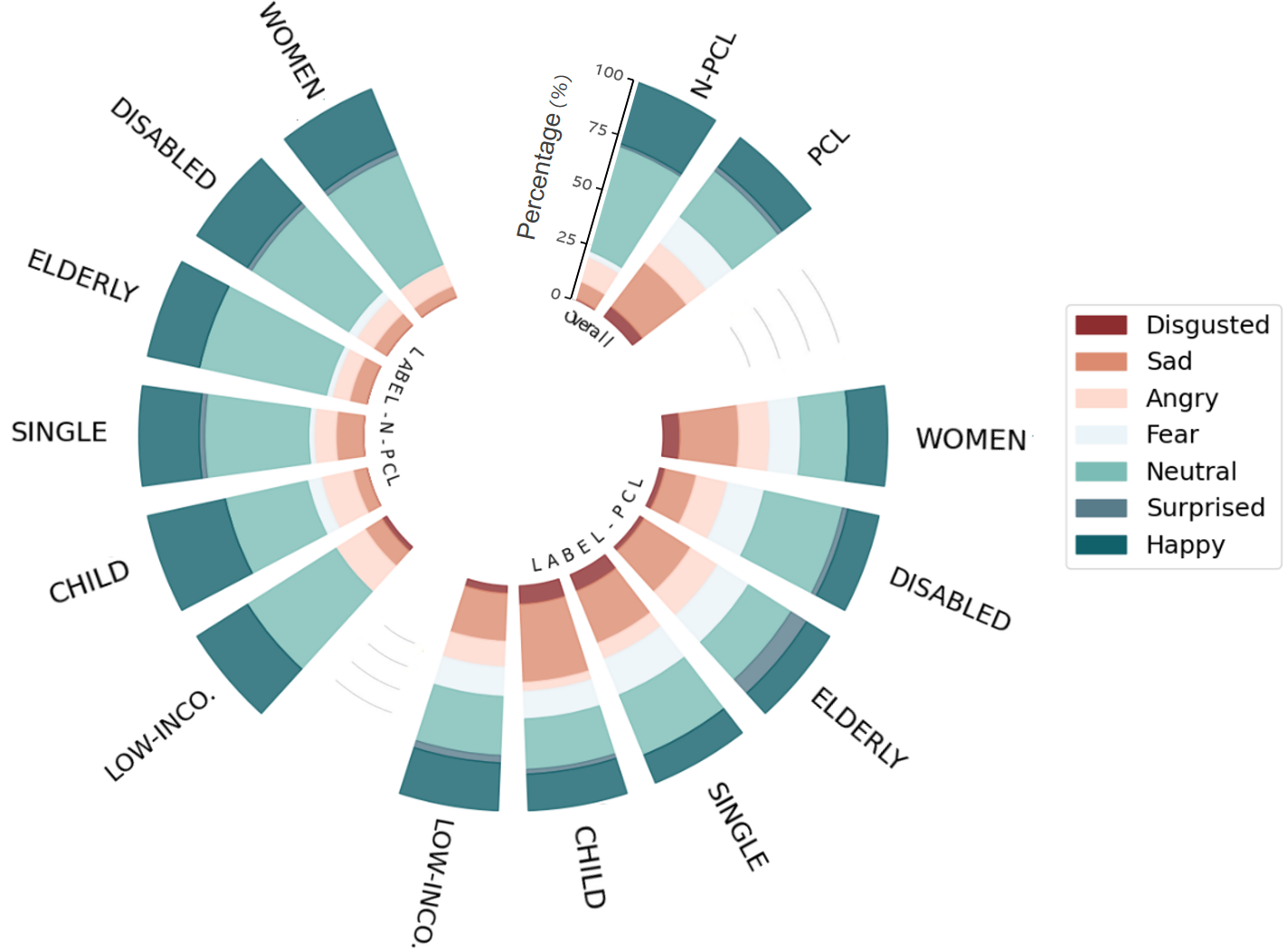} % 将图片宽度设为单栏宽度的 80%
\caption{Sentiment analysis for the six vulnerable groups in PCLMM.}
\label{fig:figure1}
\end{figure}
\subsubsection{Toxicity Analysis}
\begin{figure}[h]
\centering
\includegraphics[width=0.8\columnwidth]{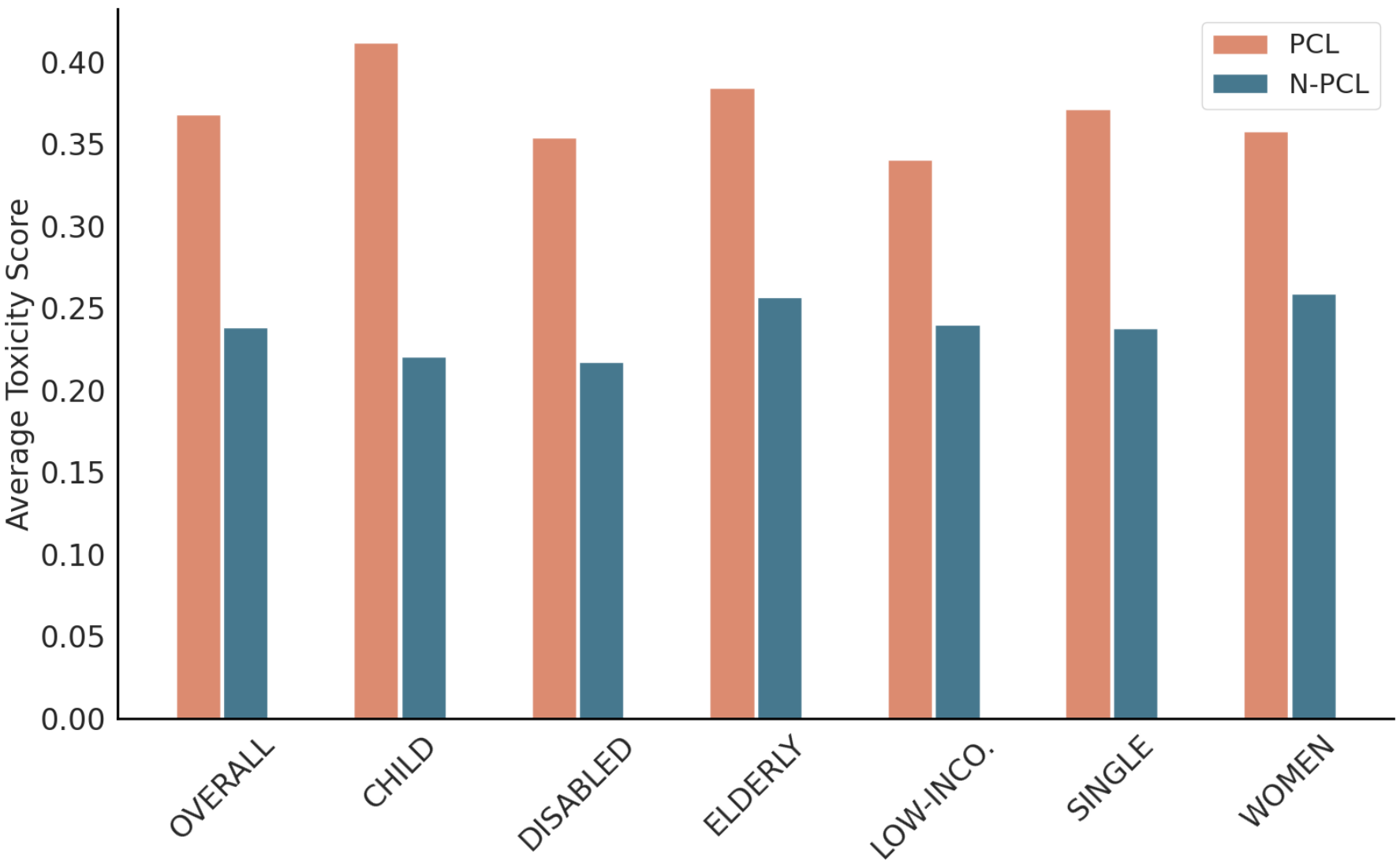} % 将图片宽度设为单栏宽度
\caption{Average toxicity scores in PCLMM.}
\label{fig:figure2}
\end{figure}

We scored our transcribed texts using the Perspective API \citep{PerspectiveAPI}, as shown in Figure \ref{fig:figure2}. PCL samples have higher toxicity scores across all community categories compared to non-PCL samples (0.37 vs. 0.24). However, PCL toxicity is lower than traditional hate speech (usually above 0.7), highlighting its implicit nature and the challenge in detection.
\section{Method}
\subsection{Problem Statement}
Given a dataset of video samples \( V \), the task is to classify videos targeting vulnerable groups as either PCL (\( y = 1 \)) or non-PCL (\( y = 0 \)). Each video \( V \) is represented by a sequence of frames \( F = \{f_1, f_2, \dots, f_n\} \) and a subset of facial expression frames \( F_v = \{f_{1v}, f_{2v}, \dots, f_{nv}\} \). If a frame \( f_n \) lacks facial expressions, \( f_{nv} \) is filled with a zero vector. The audio sequence is \( A = \{a_1, a_2, \dots, a_l\} \), and the transcribed text sequence is \( T = \{w_1, w_2, \dots, w_m\} \). The goal is to develop an attention-based multimodal classifier \( X: X(F; F_v; A; T) \rightarrow y \), where \( y \in \{0, 1\} \).

\subsection{Video Encoding}
We used the Vision Transformer (ViT) \citep{dosovitskiy2020image} to extract features from videos. Given a sequence of frames \( F = \{f_1, f_2, \dots, f_n\} \), ViT extracted feature vectors for each frame \( f_i \). The feature vector \( \mathbf{z}_i \) is computed as:
\begin{equation}
\mathbf{z}_i = \text{ViT}(f_i), \quad \mathbf{z}_i \in \mathbb{R}^{d_v}, \quad i = 1, 2, \dots, n
\end{equation}
Here, \( \mathbf{z}_i \in \mathbb{R}^{d_v} \) is the \( d_v \)-dimensional feature vector encoded by ViT for each frame \( f_i \).
\subsection{Facial Expression Encoding}
To capture patronizing facial expressions in the video, we first used MTCNN (Multi-task Cascaded Convolutional Networks) \citep{zhang2016joint} for face detection. Next, FER-VT (Facial Expression Recognition using Vision Transformers) \citep{huang2021facial} encoded facial features using grid-wise attention and visual transformers to capture long-range dependencies. For each video frame \( f_i \), if MTCNN detected a face, FER-VT extracted the facial feature vector \( \mathbf{z}_i^v \); otherwise, a zero vector was assigned.
\begin{equation}
\mathbf{z}_i^v 
\begin{cases} 
\text{FER-VT}(f_iv), & \text{if MTCNN detects a face in } f_i \\
f_iv = 0, & \text{if no face is detected}
\end{cases}
\end{equation}
\subsection{Audio Encoding}
We used FFmpeg \citep{ffmpeg}, a widely-used multimedia package, to extract high-quality audio from the videos, and then applied the Mel Frequency Cepstral Coefficient (MFCC) \citep{xu2004mfcc} to extract audio features. The extracted audio sequence \( A \) was encoded as \( \mathbf{z}^a \).
\subsection{Text Encoding}
We used Whisper \citep{radford2023whisper}, a speech recognition model by OpenAI, to transcribe audio into text. For text encoding, we utilized RoBERTa-Chinese \citep{cui2020revisiting} and a fine-tuned RoBERTa trained on the CCPC dataset \citep{wang2023ccpc} for patronizing language detection (We call it BERT-PCL). These models extract the CLS token from each transcript, producing a feature vector \( \mathbf{z}^t \).
\subsection{Cross-Modality Fusion}
In our model, we used a unified Cross-Modality Multi-Head Attention (MHCA) mechanism to fuse information across different modalities. The general form of MHCA is:
\begin{equation}
\text{MHCA}(\mathbf{Q}_i, \mathbf{K}_j, \mathbf{V}_j) = \text{Softmax}\left(\frac{\mathbf{Q}_i \mathbf{K}_j^\top}{\sqrt{d_k}}\right) \mathbf{V}_j
\end{equation}
Here, \(\mathbf{Q}_i\) is the query from modality \(i\), and \(\mathbf{K}_j\) and \(\mathbf{V}_j\) are the key and value from modality \(j\). By varying \(i\) and \(j\), the interaction between different modality pairs is expressed as:
\begin{equation}
\mathbf{A}_{i,j} = \text{MHCA}(\mathbf{Q}_i, \mathbf{K}_j, \mathbf{V}_j), \quad i,j \in \{\mathbf{z}, \mathbf{z}^v, \mathbf{z}^a, \mathbf{z}^t\}
\end{equation}
The resulting attention features are then aggregated into a unified multimodal representation:
\begin{equation}
\mathbf{Z} = \sum_{i,j} \mathbf{A}_{i,j}
\end{equation}
\subsection{Loss Function}
We employed the BCEWithLogitsLoss as our loss function, which is suitable for binary classification tasks. The loss is computed as:
\begin{equation}
\text{Loss} = -\frac{1}{N} \sum_{i=1}^{N} \left[y_i \log \sigma(x_i) + (1-y_i) \log (1-\sigma(x_i))\right]
\end{equation}
%where \( \sigma(x_i) \) is the sigmoid of the model's raw output \( x_i \), and \( y_i \) is the true label.
\section{Experiment}
\begin{table}[H]
\caption{Model performance on the classification task of PCL videos. $\mathbf{X}_{p}$ represents metrics for PCL samples. $\mathbf{F1}_{m}$ denotes the macro-averaged F1 score. Abbreviations: MC (MFCC), RC (RoBERTa-Chinese), BP (BERT-PCL), FT (FER-VT), VM (VideoMAE), VT (ViT).}
\label{tab2}
\begin{center}
\begin{tabular}{c|l|c|c|c|c|c} % Specify the correct number of columns
\toprule[1.2pt]  % Thicker top line
\textbf{M} & \textbf{Model} & \(\textbf{P}_{p}\) & \(\textbf{R}_{p}\) & \(\textbf{F1}_{p}\) & \(\textbf{F1}_{m}\) & \textbf{Acc} \\ % Header with 6 columns
\midrule
\midrule
A & MC & 35.81 & 56.89 & 45.21 & 54.28 & 64.14 \\ % Correct number of columns
\midrule
\multirow{3}{*}{T} & RC &54.84  & 50.00 & 52.31 & 69.14&78.32 \\
& BP & 58.06 & 52.94 & 55.38 & 71.13 & 79.72 \\ % Correct number of columns
& GPT4 & 65.52 & 55.88& 60.32 & 74.55 & 82.52 \\ % Correct number of columns
\midrule
F & FT & 65.52 & 47.50 & 55.07 & 70.46 & 78.47 \\ % Correct number of columns
\midrule
\multirow{2}{*}{V}& VM & 61.76 & 52.50 & 56.76 & 70.90 & 77.78  \\ % Correct number of columns
 & VT & 65.62 & 52.50 & 58.33 & 72.22 & 79.17\\ % Correct number of columns
\midrule
\midrule
A+F& MC+FT & 39.13 & 45.00 & 41.86 & 58.55 & 65.28 \\ % Correct number of columns
\midrule
A+T& MC+BP & 58.82 & 50.00 & 54.05 & 69.08 & 76.39 \\ % Correct number of columns
\midrule
T+F& BP+FT & 62.89 & 55.00 & 58.67 & 72.06 & 78.47 \\ % Correct number of columns
\midrule
A+V & MC+VT & 58.00 & 72.50 & 64.44 & 74.14 & 77.78 \\ % Correct number of columns
\midrule
V+F& VT+FT & 62.79 & 67.50 & 65.06 & 75.46 & 79.86 \\ % Correct number of columns
\midrule
V+T& VT+BP & 63.04 & 72.50 & 67.44 & 76.79 & 80.56 \\ % Correct number of columns
\midrule
\midrule
A+T+F & MC+BP+FT & 61.90 & 65.00 & 63.41 &74.43 & 79.17 \\ % Correct number of columns
\midrule
V+T+F & VT+BP+FT  & 64.44 & 72.50 & 68.24 & 77.47 & 81.25\\ % Correct number of columns
\midrule
V+T+A & VT+BP+MC & 65.91 &72.50 & 69.05 & 78.15 & 81.94 \\ % Correct number of columns
\midrule
V+A+F & VT+MC+FT & 67.44 & 72.50 & 69.88 & 78.84& 82.64 \\ % Correct number of columns
\midrule
\midrule
V+A+T+F& \textbf{MultiPCL} & \textbf{68.09} & \textbf{80.00} & \textbf{73.56} & \textbf{81.06} & \textbf{84.03} \\ % Correct number of columns
\bottomrule[1.2pt]  % Thicker bottom line
\end{tabular}
\end{center}
\end{table}
\subsection{Experimental Settings}
Our experiments were conducted using two NVIDIA A800-80G GPUs with 5-fold cross-validation to ensure robust training. We trained for 20 epochs, averaged the top five performances, and used a batch size of 10 with a learning rate of 1e-4. All code was implemented in PyTorch. Evaluation metrics included Precision, Recall, F1-score, and Accuracy, standard in toxicity detection. Notably, the ViT architecture provides an efficient solution, making it ideal for multimodal models. Its performance in short video analysis matches that of VideoMAE \citep{tong2022videomae}, which is why we chose ViT as the baseline instead of VideoMAE for modality fusion.
\subsection{Experimental Result}
We employed a strategy of progressively integrating multiple modalities, beginning with single modalities. The experimental results are presented in Table \ref{tab2}. (1) In single-modality scenarios, the text modality yielded the highest detection performance, while using audio alone resulted in poor outcomes, underscoring the ongoing importance of text in toxicity detection. (2) In multi-modality scenarios, incorporating the video modality often leads to superior results. For dual-modality setups, combinations that include video achieved an average F1 score of 75.46, whereas those without video only reached 66.56. This trend is also evident in tri-modal configurations, highlighting the critical supportive role of video in feature understanding. Moreover, the facial expression modality only demonstrates optimal performance when combined with the video modality. (3) Our proposed MultiPCL, which integrates four modalities, significantly outperforms all baselines, with performance improvements of 6.51\%, 4.27\%, and 2.22\% over the best single, dual, and tri-modal setups, respectively. This confirms the effectiveness of our detector.

We further conducted ablation experiments (Table \ref{tab3}) on the MultiPCL detector to demonstrate the role of MHCA. Our experiments showed that replacing MHCA with a standard fully connected layer resulted in nearly a 4\% decrease in F1 score, highlighting the critical role of MHCA in capturing the relationships between different modalities.
\begin{table}[H]
\caption{Ablation Study to show the effectiveness of MHCA.}
\label{tab3}
\begin{center}
\begin{tabular}{l|c|c|c|c|c} % Specify the correct number of columns
\toprule[1.2pt]  % Thicker top line
 \textbf{Model} & \(\textbf{P}_{p}\) & \(\textbf{R}_{p}\) & \(\textbf{F1}_{p}\) & \(\textbf{F1}_{m}\) & \textbf{Acc} \\ % Header with 6 columns
\midrule
 \textbf{MultiPCL} & \textbf{68.09} & \textbf{80.00} & \textbf{73.56} & \textbf{81.06} & \textbf{84.03} \\ % Correct number of columns
-MHCA &62.50 & 75.00 & 68.18 & 77.09 & 80.56 \\ % Correct number of columns
\bottomrule[1.2pt]  % Thicker bottom line
\end{tabular}
\end{center}
\end{table}
\section{Conclusion}
Patronizing and Condescending Language (PCL) is a form of discriminatory speech targeting vulnerable groups and is widespread online, demanding more comprehensive data resources and detection schemes. In this paper, we present PCLMM, the first multimodal PCL video dataset with 715 annotated videos totaling over 21 hours. We also propose the MultiPCL detector, integrating video and discriminatory facial expression features for multimodal detection, achieving state-of-the-art performance. Future work will explore PCL's impact on microaggressions such as sarcasm and stereotypes, and evaluate existing multimodal large language models, particularly those incorporating audio, using our dataset and detector as benchmarks for microaggression detection.
\bibliographystyle{apalike} % 选择引用样式，如 plain, IEEEtran, unsrt 等
\bibliography{ref} % 这里是你的.bib文件名，不需要扩展名

\vspace{12pt}

\end{document}